\documentclass[11pt,a4paper,dvipsnames]{article}
\usepackage[hyperref]{acl2019}

\usepackage{times}
\usepackage{latexsym}
\usepackage{algorithm}
\usepackage[noend]{algpseudocode}
\usepackage{amsmath}

\usepackage{subfigure}

\usepackage{epsfig}
\usepackage{graphicx}
\usepackage{amsmath}
\usepackage{amssymb}
\usepackage{color}
\usepackage{kky}
\usepackage{multirow}
\usepackage{array}
\usepackage{booktabs}
\usepackage{xcolor,colortbl}
\definecolor{lightslategray}{rgb}{0.9, 0.9, 0.9}
\usepackage[normalem]{ulem}

\newcolumntype{L}{>{\raggedleft\arraybackslash}p{0.04in}}
\newcolumntype{N}{>{\arraybackslash}p{1.55 in}}
\newcolumntype{M}{>{\centering\arraybackslash}m{.5in}}
\newcolumntype{H}{>{\centering\arraybackslash}m{1.2in}}

\usepackage{url}

\aclfinalcopy % Uncomment this line for the final submission

\title{Domain Adaptation of Neural Machine Translation by Lexicon Induction}

\author{Junjie Hu, Mengzhou Xia, Graham Neubig, Jaime Carbonell \\
  Language Technologies Institute \\
  School of Computer Science \\
  Carnegie Mellon University \\
  {\tt \{junjieh,gneubig,jgc\}@cs.cmu.edu, mengzhox@andrew.cmu.edu}\\}

\date{}
\begin{document}
\maketitle
\begin{abstract}
  It has been previously noted that neural machine translation (NMT) is very sensitive to domain shift. In this paper, we argue that this is a dual effect of the highly lexicalized nature of NMT, resulting in failure for sentences with large numbers of unknown words, and lack of supervision for domain-specific words.  To remedy this problem, we propose an unsupervised adaptation method which fine-tunes a pre-trained out-of-domain NMT model using a pseudo-in-domain corpus. Specifically, we perform lexicon induction to extract an in-domain lexicon, and construct a pseudo-parallel in-domain corpus by performing word-for-word back-translation of monolingual in-domain target sentences. In five domains over twenty pairwise adaptation settings and two model architectures, our method achieves consistent improvements without using any in-domain parallel sentences, improving up to 14 BLEU over unadapted models, and up to 2 BLEU over strong back-translation baselines. {\let\thefootnote\relax\footnote{{Code/scripts are released at \url{https://github.com/junjiehu/dali}.}}}
  %\footnote{Code to replicate experiments will be released upon acceptance. }  \jh{hide the superscript of the footnote in case there is a mismatch with the one before abstract due}
\end{abstract}

\begin{figure*}[th]
  \centering
  \includegraphics[width=0.9\linewidth]{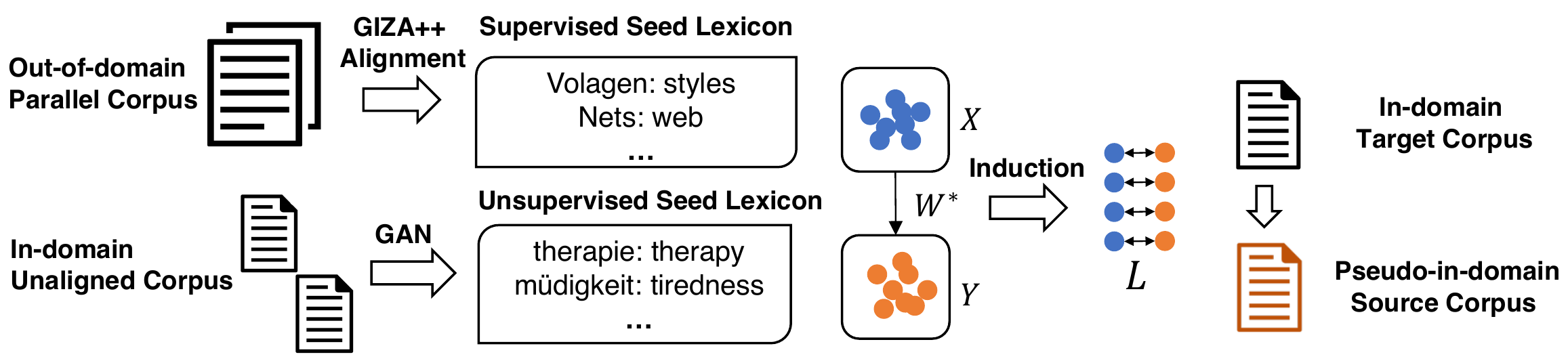}
%   \vspace{-1mm}
  \caption{Work flow of domain adaptation by lexicon induction (DALI).}
%   \vspace{-4mm}
  \label{fig:word-level-1}
\end{figure*}

\section{Introduction}

%\gn{It might be nice to have a figure on the top right of the first page illustrating the proposed method.}
%\jh{Thanks! we will make a figure}

Neural machine translation (NMT) has demonstrated impressive performance when trained on large-scale corpora \cite{bojar-EtAl:2018:WMT1}.
However, it has also been noted that NMT models trained on corpora in a particular domain tend to perform poorly when translating sentences in a significantly different domain \cite{chu-wang:2018:C18-1, koehn-knowles:2017:NMT}.
Previous work in the context of phrase-based statistical machine translation~\cite{daume2011domain} has noted that unseen (OOV) words account for a large portion of translation errors when switching to new domains.
However this problem of OOV words in cross-domain transfer is under-examined in the context of NMT, where both training methods and experimental results will differ greatly.
In this paper, we try to fill this gap, examining domain adaptation methods for NMT specifically focusing on correctly translating unknown words.

As noted by \newcite{chu-wang:2018:C18-1}, there are two important distinctions to make in adaptation methods for MT.
The first is data requirements; \emph{supervised} adaptation relies on in-domain parallel data, and \emph{unsupervised} adaptation has no such requirement.
There is also a distinction between  \emph{model-based} and \emph{data-based} methods.
Model-based methods make explicit changes to the model architecture such as jointly learning domain discrimination and translation \cite{W17-4712}, interpolation of language modeling and translation \cite{gulcehre2015using, domhan-hieber:2017:EMNLP2017}, and domain control by adding tags and word features \cite{R17-1049}. On the other hand, data-based methods perform adaptation either by combining in-domain and out-of-domain parallel corpora for supervised adaptation~\cite{luong2015stanford,freitag2016fast} or by generating pseudo-parallel corpora from in-domain monolingual data for unsupervised adaptation~\cite{sennrich-haddow-birch:2016:P16-11,currey-micelibarone-heafield:2017:WMT}. 

Specifically, in this paper we tackle the task of \emph{data-based, unsupervised} adaptation, where representative methods include creation of a pseudo-parallel corpus by back-translation of in-domain monolingual target sentences~\cite{sennrich-haddow-birch:2016:P16-11}, or construction of a pseudo-parallel in-domain corpus by copying monolingual target sentences to the source side~\cite{currey-micelibarone-heafield:2017:WMT}.
However, while these methods have potential to strengthen the target-language decoder through addition of in-domain target data, they do not explicitly provide direct supervision of domain-specific words, which we argue is one of the major difficulties caused by domain shift.

To remedy this problem, we propose a new data-based method for unsupervised adaptation that specifically focuses the unknown word problem: \textbf{domain adaptation by lexicon induction (DALI)}. Our proposed method leverages large amounts of monolingual data to find translations of in-domain unseen words, and constructs a pseudo-parallel in-domain corpus via word-for-word back-translation of monolingual in-domain target sentences into source sentences. More specifically, we leverage existing supervised \cite{xing-EtAl:2015:NAACL-HLT} and unsupervised \cite{conneau2017word} lexicon induction methods that project source word embeddings to the target embedding space, and find translations of unseen words by their nearest neighbors. For supervised lexicon induction, we learn such a mapping function under the supervision of a seed lexicon extracted from out-of-domain parallel sentences using word alignment. For unsupervised lexicon induction, we follow \newcite{conneau2017word} to infer a lexicon by adversarial training and iterative refinement.

In the experiments on German-to-English translation across five domains (Medical, IT, Law, Subtitles, and Koran), we find that DALI improves both RNN-based \cite{bahdanau2014neural} and Transformer-based \cite{vaswani2017attention} models trained on an out-of-domain corpus with gains as high as 14 BLEU. When the proposed method is combined with back-translation, we can further improve performance by up to 4 BLEU. Further analysis shows that the areas in which gains are observed are largely orthogonal to back-translation; our method is effective in translating in-domain unseen words, while back-translation mainly improves the fluency of source sentences, which helps the training of the NMT decoder. 

%\gn{Nice intro! I'm happy up to here so I won't be re-checking the intro unless necessary.}

\section{Domain Adaptation by Lexicon Induction}
Our method works in two steps: (1) we use lexicon induction methods to learn an in-domain lexicon from in-domain monolingual source data $D_\text{src-in}$ and target data $D_\text{tgt-in}$ as well as out-of-domain parallel data $D_\text{parallel-out}$, (2) we use this lexicon to create a pseudo-parallel corpus for MT.
% In the following sections, we describe both of these steps in turn.
% We use boldfaced capital characters to denote matrices, e.g., $\Xb$, $\Yb$, $\Wb$, and boldfaced lowercased characters to denote vectors, e.g., $\xb$, $\yb$. \gn{I commented this out, as it's probably not necessary and somewhat breaks the flow. It could be re-added though.}

\subsection{Lexicon Induction}

Given separate source and target word embeddings, $\Xb$, $\Yb\in \mathbb{R}^{d\times N}$, trained on all available monolingual source and target sentences across all domains, we leverage existing lexicon induction methods that perform \emph{supervised}~\cite{xing-EtAl:2015:NAACL-HLT} or \emph{unsupervised}~\cite{conneau2017word} learning of a mapping $f(\Xb)=\Wb\Xb$ that transforms source embeddings to the target space, then selects nearest neighbors in embedding space to extract translation lexicons. 

\paragraph{Supervised Embedding Mapping}
% \gn{You need to explain the seed lexicon requirements first, and make sure the use of the seed lexicon is reflected in your equations.}
% \gn{Move this sentence earlier.}.
Supervised learning of the mapping function requires a seed lexicon of size $n$, denoted as $L=\{(s,t)_i\}_{i=1}^n$. We represent the source and target word embeddings of the $i$-th translation pair $(s,t)_i$ by the $i$-th column vectors of $\Xb^{(n)}, \Yb^{(n)} \in \mathbb{R}^{d\times n}$ respectively.  \newcite{xing-EtAl:2015:NAACL-HLT} show that by enforcing an orthogonality constraint on $\Wb\in O_d(\mathbb{R})$, we can obtain a closed-form solution from a singular value decomposition (SVD) of $\Yb^{(n)} {\Xb^{(n)}}^T$:

\begin{align}
    &\Wb^* = \arg \max_{\Wb \in O_d(\mathbb{R})} \|\Yb^{(n)} - \Wb \Xb^{(n)}\|_F = \Ub \Vb^T \nonumber \\ \label{eq:svd}
    &\Ub \Sigma \Vb^T = \text{SVD}(\Yb^{(n)} {\Xb^{(n)}}^T).
\end{align}

%\gn{Hmm, I took a look at this and it's strange. It looks like you're not actually using the GIZA++ alignments themselves, but rather using GIZA++ to filter the lexicon, then calculating co-occurrence probabilities directly. We should discuss.}
%The next question is how to create a seed lexicon that can bootstrap this process.
In a domain adaptation setting we have parallel out-of-domain data $D_\text{parallel-out}$, which can be used to extract a seed lexicon. Algorithm~\ref{alg:seed} shows the procedure of extracting this lexicon. We use the word alignment toolkit GIZA++ \cite{och03:asc} to extract word translation probabilities $P(t|s)$ and $P(s|t)$ in both forward and backward directions from $D_\text{parallel-out}$, and extract lexicons $L_\text{fw}=\{(s,t),~\forall P(t|s)>0\}$ and  $L_\text{bw}=\{(s,t),~\forall P(s|t)>0\}$. We take the union of the lexicons in both directions and further prune out translation pairs containing punctuation that is non-identical. To avoid multiple translations of either a source or target word, we find the most common translation pairs in $D_\text{parallel-out}$, sorting translation pairs by the number of times they occur in $D_\text{parallel-out}$ in descending order, and keeping those pairs with highest frequency in $D_\text{parallel-out}$.

% \jh{Thanks for the suggestions! Yes. We did such an experiment in Table~\ref{tab:comparison}}

\begin{algorithm}[t]
\caption{Supervised lexicon extraction}\label{alg:seed}
    \hspace*{\algorithmicindent} \textbf{Input}: Parallel out-of-domain data $D_\text{parallel-out}$ \\
    \hspace*{\algorithmicindent} \textbf{Output}: Seed lexicon $L=\{(s,t)\}_{i=1}^n$
\begin{algorithmic}[1]
% \Procedure{Train}{}
\State Run GIZA++ on $D_\text{parallel-out}$ to get $L_\text{fw}, L_\text{bw}$
\State $L_g = L_\text{fw} \cup L_\text{bw}$
\State Remove pairs with punctuation only in either $s$ and $t$ from $L_g$
\State Initialize a counter $C[(s, t)]=0~\forall (s, t) \in L_g$
\For {(src, tgt) $\in D_\text{parallel-out}$}
    \For {$(s, t) \in L_g$}
        \If {$s\in \text{src}$ \textbf{and} $t\in \text{tgt}$ } 
            \State $C[(s, t)] = C[(s, t)] + 1$
        \EndIf
    \EndFor
\EndFor
\State Sort $C$ by its values in the descending order 
\State $L = \{\}, S=\{\}, T=\{\}$ 
\For {$(s, t) \in C$}
    \If {$s\notin S$ \textbf{and} $t\notin T$ }
        \State $L=L\cup \{(s, t)\}$
        \State $S=S\cup\{s\},~T=T\cup\{t\}$
    \EndIf
\EndFor
\State \textbf{return} $L$
% \EndProcedure
\end{algorithmic}
\end{algorithm}

\paragraph{Unsupervised Embedding Mapping}
For unsupervised training, we follow \newcite{conneau2017word} in mapping source word embeddings to the target word embedding space through adversarial training.
Details can be found in the reference, but briefly a discriminator is trained to distinguish between an embedding sampled from $\Wb\Xb$ and $\Yb$, and $\Wb$ is trained to prevent the discriminator from identifying the origin of an embedding by making $\Wb\Xb$ and $\Yb$ as close as possible.

\paragraph{Induction} Once we obtain the matrix $\Wb$ either from supervised or unsupervised training, we map all the possible in-domain source words to the target embedding space. We compute the nearest neighbors of an embedding by a distance metric, Cross-Domain Similarity Local Scaling (CSLS; \citet{conneau2017word}):  %We consider a bi-partite neighborhood graph, in which each word of a given dictionary is connected to its $K$ nearest neighbors in the other languages. We denote by $\mathcal{N}_T(h_s)$ the neighborhood, on this bi-partite graph, associated with a source word representation $h_s$. 

\begin{align} \label{eq:csls} \nonumber
    &\text{CSLS}(\Wb\xb, \yb) = 2\cos(\Wb \xb, \yb)  - r_T(\Wb \xb)- r_S(\yb)  \\ \nonumber
    &r_T(\Wb\xb) = {1\over K}\sum_{\yb' \in \Ncal_T(\Wb\xb)}\cos(\Wb \xb,\yb')
\end{align}where $r_T(\Wb\xb)$ and $r_S(\yb)$ measure the average cosine similarity between their $K$ nearest neighbors in the source and target spaces respectively.

To ensure the quality of the extracted lexicons, we only consider mutual nearest neighbors, i.e., pairs of words that are mutually nearest neighbors of each other according to CSLS. This significantly decreases the size of the extracted lexicon, but improves the reliability. 

%\jgc{Any idea how large the seed bilingual lexicon needs to be?}
%\jh{Previous work finds that using a seed dictionary of a few thousand pairs can learn a good mapping when the language pairs are very similar.}

\subsection{NMT Data Generation and Training}

Finally, we use this lexicon to create pseudo-parallel in-domain data to train NMT models.
Specifically, we follow \newcite{sennrich-haddow-birch:2016:P16-11} in back-translating the in-domain monolingual target sentences to the source language, but instead of using a pre-trained target-to-source NMT system, we simply perform word-for-word translation using the induced lexicon $L$. Each target word in the target side of $L$ can be deterministically back-translated to a source word, since we take the nearest neighbor of a target word as its translation according to CSLS. If a target word is not mutually nearest to any source word, we cannot find a translation in $L$ and we simply copy this target word to the source side. We find that more than 80\% of the words can be translated by the induced lexicons. We denote the constructed pseudo-parallel in-domain corpus as $D_\text{pseudo-parallel-in}$.
% \gn{But you said that you only take mutual nearest neighbors in the previous section... If a target is not a mutual neighbor with the source word what happens?}.

During training, we first pre-train an NMT system on an out-of-domain parallel corpus $D_\text{parallel-out}$, and then fine tune the NMT model on a constructed parallel corpus.
More specifically, to avoid overfitting to the extracted lexicons, we sample an equal number of sentences from $D_\text{parallel-out}$, and get a fixed subset $D'_\text{parallel-out}$, where $|D'_\text{parallel-out}|=|D_\text{pseudo-parallel-in}|$.  We concatenate $D'_\text{parallel-out}$ with $D_\text{pseudo-parallel-in}$, and fine-tune the NMT model on the combined corpus.  % \gn{Do you mean you ``sample an equal number of sentences from $D_\text{parallel-out}$ and $D_\text{pseudo-parallel-in}$ in each batch during fine-tuning''?}.

\begin{table*}[t]
\centering
\begin{tabular}{c|l|l||r|r|r|r|r||r|l}
\hline
Domain                     & \multicolumn{2}{c||}{Method}              & Medical & IT    & Subtitles & Law   & Koran & \multicolumn{1}{c|}{Avg.} & \multicolumn{1}{c}{Gain} \\ \hline \hline
\multirow{4}{*}{Medical}   & \multirow{2}{*}{LSTM}        & Unadapted & \cellcolor{lightslategray}{46.19}   & 4.62  & 2.54      & 7.05  & 1.25  & 3.87                           & \multirow{2}{*}{+4.31}          \\ \cline{3-9}
                           &                              & DALI   & \cellcolor{lightslategray}{-}       & 11.32 & 7.79      & 9.72  & 3.85  & \bf{8.17}                           &                                \\ \cline{2-10} 
                           & \multirow{2}{*}{XFMR} & Unadapted & \cellcolor{lightslategray}{49.66}   & 4.54  & 2.39      & 7.77  & 0.93  & 3.91                           & \multirow{2}{*}{+4.79}          \\ \cline{3-9}
                           &                              & DALI   & \cellcolor{lightslategray}{-}       & 10.99 & 8.25      & 11.32 & 4.22  & \bf{8.70}                           &                                \\ \hline \hline
\multirow{4}{*}{IT}        & \multirow{2}{*}{LSTM}        & Unadapted & 7.43    & \cellcolor{lightslategray}{57.79} & 5.49      & 4.10  & 2.52  & 4.89                           & \multirow{2}{*}{+5.98}          \\ \cline{3-9}
                           &                              & DALI   & 20.44   & \cellcolor{lightslategray}{-}     & 9.53      & 8.63  & 4.85  & \bf{10.86}                          &                                \\ \cline{2-10} 
                           & \multirow{2}{*}{XFMR} & Unadapted & 6.96    & \cellcolor{lightslategray}{60.43} & 6.42      & 4.50  & 2.45  & 5.08                           & \multirow{2}{*}{+5.76}          \\ \cline{3-9}
                           &                              & DALI   & 19.49   & \cellcolor{lightslategray}{-}     & 10.49     & 8.75  & 4.62  & \bf{10.84}                          &                                \\ \hline \hline
\multirow{4}{*}{Subtitles} & \multirow{2}{*}{LSTM}        & Unadapted & 11.36   & 12.27 & \cellcolor{lightslategray}{27.29}     & 10.95 & 10.57 & 11.29                          & \multirow{2}{*}{+2.79}          \\ \cline{3-9}
                           &                              & DALI   & 21.63   &  12.99   & \cellcolor{lightslategray}{-}         & 11.50 &   10.17    & \bf{16.57}                          &                                \\ \cline{2-10} 
                           & \multirow{2}{*}{XFMR} & Unadapted & 16.51   & 14.46 & \cellcolor{lightslategray}{30.71}     & 11.55 & 12.96 & 13.87                          & \multirow{2}{*}{+3.85}          \\ \cline{3-9}
                           &                              & DALI   & 26.17   & 17.56 & \cellcolor{lightslategray}{-}         & 13.96 & 13.18 & \bf{17.72}                          &                                \\ \hline \hline
\multirow{4}{*}{Law}       & \multirow{2}{*}{LSTM}        & Unadapted & 15.91   & 6.28  & 4.52      & \cellcolor{lightslategray}{40.52} & 2.37  & 7.27                           & \multirow{2}{*}{+4.85}          \\ \cline{3-9}
                           &                              & DALI   & 24.57   & 10.07 & 9.11      & \cellcolor{lightslategray}{-}     & 4.72  & \bf{12.12}                          &                                \\ \cline{2-10} 
                           & \multirow{2}{*}{XFMR} & Unadapted & 16.35   & 5.52  & 4.57      & \cellcolor{lightslategray}{46.59} & 1.82  & 7.07                           & \multirow{2}{*}{+6.17}          \\ \cline{3-9}
                           &                              & DALI   & 26.98   & 11.65 & 9.14      & \cellcolor{lightslategray}{-}     & 5.15  & \bf{13.23}                          &                                \\ \hline \hline
\multirow{4}{*}{Koran}     & \multirow{2}{*}{LSTM}        & Unadapted & 0.63    & 0.45  & 2.47      & 0.67  & \cellcolor{lightslategray}{19.40} & 1.06                           & \multirow{2}{*}{+6.56}          \\ \cline{3-9}
                           &                              & DALI   & 12.90   & 5.25  & 7.49      & 4.80  & \cellcolor{lightslategray}{-}     & \bf{7.61}                           &                                \\ \cline{2-10} 
                           & \multirow{2}{*}{XFMR} & Unadapted & 0.00    & 0.44  & 2.58      & 0.29  & \cellcolor{lightslategray}{15.53} & 0.83                           & \multirow{2}{*}{+7.54}          \\ \cline{3-9}
                           &                              & DALI   & 14.27   & 5.24  & 9.01      & 4.94  & \cellcolor{lightslategray}{-}     & \bf{8.37}                           &                                \\ \hline
\end{tabular}
% \vspace{-1mm}
\caption{BLEU scores of LSTM based and Transformer (XFMR) based NMT models when trained on one domain (columns), and tested on another domain (rows). The last two columns show the average performance of unadapted baselines and DALI, and the average gains.} \label{tab:main} 
% \vspace{-3mm}
\end{table*}

\section{Experimental Results}
\subsection{Data}

We follow the same setup and train/dev/test splits of \newcite{koehn-knowles:2017:NMT}, using a German-to-English parallel corpus that covers five different domains. %: Medical, IT, Subtitles, Law, and Koran.
Data statistics are shown in Table~\ref{tab:stats}. Note that these domains are very distant from each other. Following \newcite{koehn-knowles:2017:NMT}, we process all the data with byte-pair encoding~\cite{sennrich-haddow-birch:2016:P16-12} to construct a vocabulary of 50K subwords. To build an unaligned monolingual corpus for each domain, we randomly shuffle the parallel corpus and split the corpus into two parts with equal numbers of parallel sentences. We use the target and source sentences of the first and second halves respectively.  We combine all the unaligned monolingual source and target sentences on all five domains to train a skip-gram model using \textit{fasttext}~\cite{bojanowski2017enriching}. We obtain source and target word embeddings in 512 dimensions by running 10 epochs with a context window of 10, and 10 negative samples. 

% \subsection{Data Statistic} \label{sec:data_stat}
\begin{table}[h]
\centering
\resizebox{0.48\textwidth}{!}{%
\begin{tabular}{l||r||r||r}
\hline
Corpus         & Words     & Sentences & W/S \\ \hline
Medical & 12,867,326  & 1,094,667   & 11.76     \\ 
IT             & 2,777,136   & 333,745    & 8.32  \\
Subtitles      & 106,919,386 & 13,869,396  & 7.71  \\
Law  & 15,417,835  & 707,630    & 21.80     \\ 
Koran & 9,598,717   & 478,721    & 20.05     \\ \hline
\end{tabular} }
% \vspace{-1mm}
\caption{Corpus statistics over five domains.}
% \vspace{-3mm}
\label{tab:stats}
\end{table}

\subsection{Main Results}
% (hyper-parameter details can be found in Appendix~\ref{tab:parameters}).
We first compare DALI with other adaptation strategies on both RNN-based and Transformer-based NMT models. 

Table~\ref{tab:main} shows the performance of the two models when trained on one domain (columns) and tested on another domain (rows). We fine-tune the unadapted baselines using pseudo-parallel data created by DALI. We use the unsupervised lexicon here for all settings, and leave a comparison across lexicon creation methods to Table~\ref{tab:comparison}. Based on the last two columns in Table~\ref{tab:main}, DALI substantially improves both NMT models with average gains of 2.79-7.54 BLEU over the unadapted baselines. % across all domains. 
%\jh{I am thinking of adding a graph or table to show the distance between domains, and see if the gains can be associated with the extend of domain mismatch. For example, $distance(D_\text{parallel-out}, D_\text{parallel-in}) > distance(D_\text{pseudo-parallel-in}, D_\text{parallel-in})$. One way is to estimate the probability distribution of words by their frequency in a corpus, and compute the Jensen–Shannon divergence between two discrete distributions. Any idea on measuring the domain mismatch?}
%\gn{Another way is perplexity according a (BPE-based?) language model, which has been used in \newcite{moore2010intelligent}.}

We further compare DALI with two popular data-based unsupervised adaptation methods that leverage in-domain monolingual target sentences: (1) a method that copies target sentences to the source side (Copy; \newcite{currey-micelibarone-heafield:2017:WMT}) and (2) back-translation (BT; \newcite{sennrich-haddow-birch:2016:P16-11}), which translates target sentences to the source language using a backward NMT model. We compare DALI with supervised (DALI-S) and unsupervised (DALI-U) lexicon induction. Finally, we (1) experiment with when we directly extract a lexicon from an in-domain corpus using GIZA++ (DALI-GIZA++) and Algorithm~\ref{alg:seed}, and (2) list scores for when systems are trained directly on in-domain data (In-domain). For simplicity, we test the adaptation performance of the LSTM-based NMT model, and train a LSTM-based NMT with the same architecture on out-of-domain corpus for English-to-German back-translation.

First, DALI is competitive with BT, outperforming it on the medical domain, and underperforming it on the other three domains. Second, the gain from DALI is orthogonal to that from BT -- when combining the pseudo-parallel in-domain corpus obtained from DALI-U with that from BT, we can further improve by 2-5 BLEU points on three of four domains. Second, the gains through usage of both DALI-U and DALI-S are surprisingly similar, although the lexicons induced by these two methods have only about 50\% overlap. Detailed analysis of two lexicons can be found in Section~\ref{sec:lex_coverage}. 

\begin{table}[t]
\resizebox{0.48\textwidth}{!}{
\begin{tabular}{l|r|r|r|r}
\hline
             & Medical & Subtitles & Law & Koran \\ \hline\hline
Unadapted    & 7.43                     & 5.49                       & 4.10                 & 2.52                   \\ \hline
Copy         & 13.28                    & 6.68                       & 5.32                 & 3.22                   \\ 
BT           & 18.51                    & 11.25                      & 11.55                & \textbf{8.18}                   \\ 
DALI-U     & 20.44                    & 9.53                       & 8.63                 & 4.90                   \\ 
DALI-S       & 19.03                    & 9.80                       & 8.64                 & 4.91                   \\ 
DALI-U+BT & \textbf{24.34}                    & \textbf{13.35}                      & \textbf{13.74}                & 8.11                   \\ \hline
DALI-GIZA++      & 28.39                    &  9.37                          & 11.45                & 8.09    \\
In-domain        & 46.19                    &  27.29                          & 40.52                & 19.40    \\
\hline\hline
\end{tabular}}
% \vspace{-1mm}
\caption{Comparison among different methods on adapting NMT from IT to \{Medical, Subtitles, Law, Koran\} domains, along with two oracle results}
\label{tab:comparison}
% \vspace{-3mm}
\end{table}

\subsection{Word-level Translation Accuracy}
Since our proposed method focuses on leveraging word-for-word translation for data augmentation, we analyze the word-for-word translation accuracy for unseen in-domain words. A source word is considered as an unseen in-domain word when it never appears in the out-of-domain corpus.
We examine two question: (1) How much does each adaptation method improve the translation accuracy of unseen in-domain words? (2) How does the frequency of the in-domain word affect its translation accuracy?

To fairly compare various methods, we use a lexicon extracted from the in-domain parallel data with the GIZA++ alignment toolkit as a reference lexicon $L_g$.
% Though $L_g$ is not hand-curated, more than 80\% of the time the corresponding target word in $L_g$ also appears in the reference sentence, indicating that it is a reasonable approximation to the gold standard.
For each unseen in-domain source word in the test file, when the corresponding target word in $L_g$ occurs in the output, we consider it as a ``hit'' for the word pair. 

First, we compare the percentage of successful in-domain word translations across all adaptation methods. Specifically, we scan the source and reference of the test set to count the number of valid hits $C$, then scan the output file to get the count $C_t$ in the same way. Finally, the hit percentage is calculated as $\frac{C_t}{C}$. The results on experiments adapting IT to other domains are shown in Figure~\ref{fig:word-level-1}. The hit percentage of the unadapted output is extremely low, which confirms our assumption that in-domain word translation poses a major challenge in adaptation scenarios.   We also find that all augmentation methods can improve the translation accuracy of unseen in-domain words but our proposed method can outperform all others in most cases. The unseen in-domain word translation accuracy is quantitatively correlated with the BLEU scores, which shows that correctly translating in-domain unseen words is a major factor contributing to the improvements seen by these methods.

% \jh{Translation accuracy of in-domain corpus}
\begin{figure}[t]
  \includegraphics[width=\linewidth]{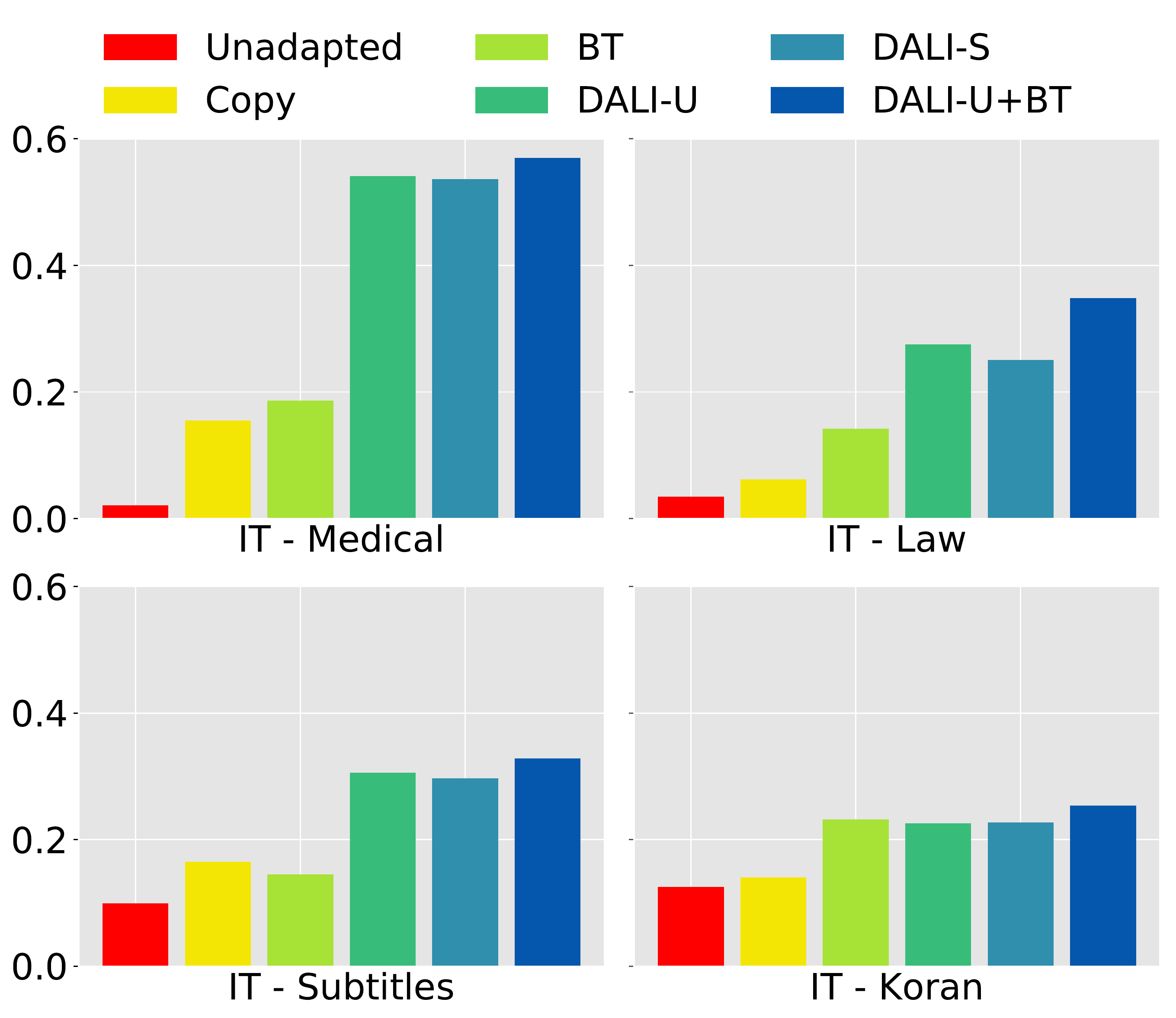}
%   \vspace{-1mm}
  \caption{Translation accuracy of in-domain words of the test set on several data augmentation baseline and our proposed method with IT as the out domain} % \gn{This is saying ``acquis''. Let's use ``U'' instead of ``US'' to be concise, including in the figure.}.}
  \label{fig:word-level-1}
%   \vspace{-6mm}
\end{figure}
% \vspace{-4mm}

\begin{figure}[t]
  \includegraphics[width=\linewidth]{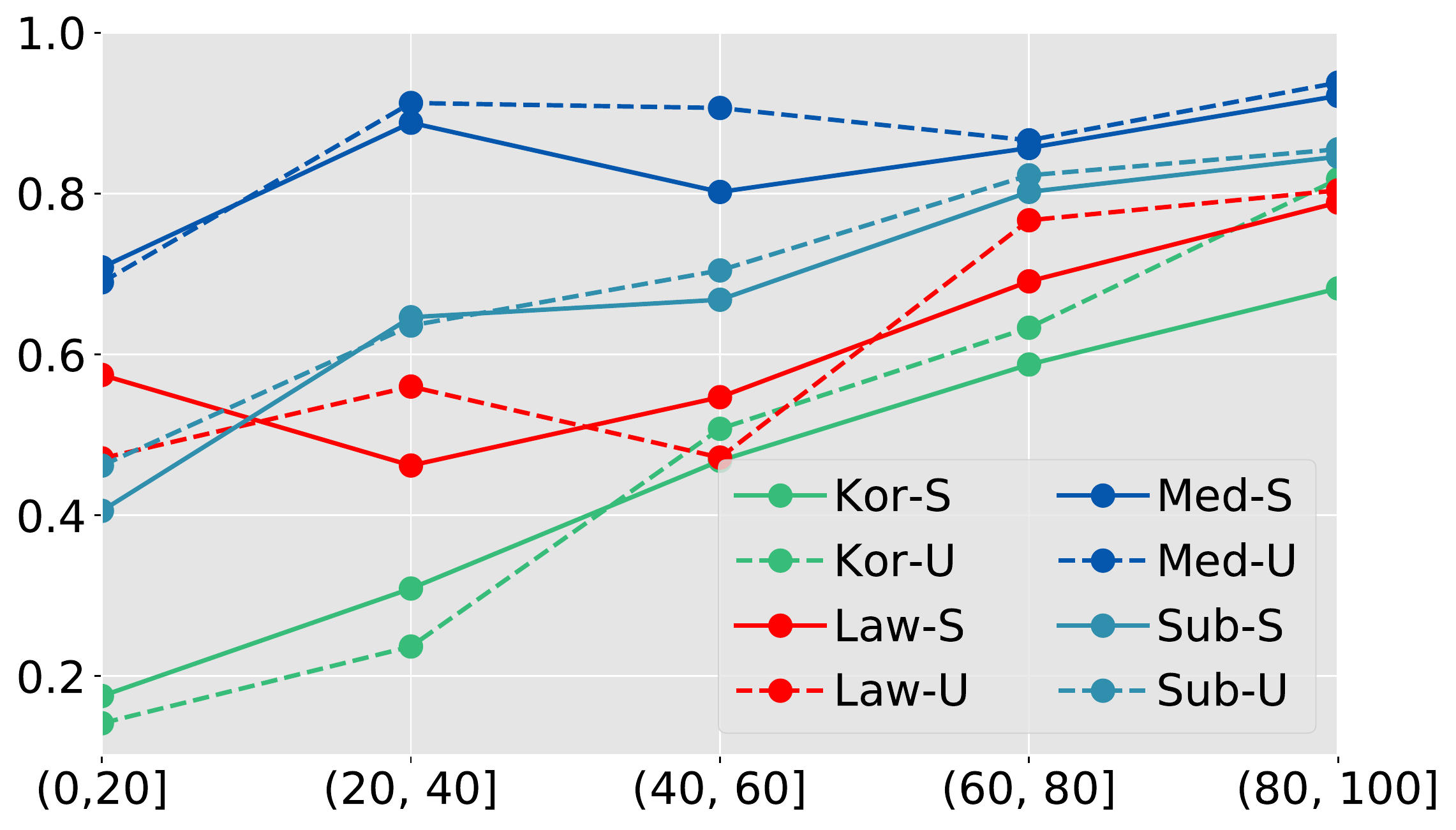}
%   \vspace{-1mm}
  \caption{Translation accuracy of in-domain unseen words in the test set with regards to the frequency percentile of lexicon words inserted in the pseudo-in-domain training corpus.}
  % \gn{I think it would probably be more interesting to see a comparison between DALI and BT instead of between DALI-U and DALI-S here.}
  \label{fig:word-level-2}
%   \vspace{-3mm}
\end{figure}

% \subsection{How does frequency of the in-domain word affect its translation accuracy?}
Second, to investigate the effect of frequency of word-for-word translation, we bucket the unseen in-domain words by their frequency percentile in the pseudo-in-domain training dataset, and calculate calculate the average translation accuracy of unseen in-domain words within each bucket. The results are plotted in Figure~\ref{fig:word-level-2} in which the x-axis represents each bucket within a range of frequency percentile, and the y-axis represents the average translation accuracy. With increasing frequency of words in the pseudo-in-domain data, the translation accuracy also increases, which is consistent with our intuition that the neural network would be able to remember high frequency tokens better. Since the absolute value of the occurrences are different among all domains, the numerical values of accuracy within each bucket vary across domains, but all lines follow the ascending pattern.

\begin{table*}[t]
\centering
\resizebox{0.95\textwidth}{!}{
\begin{tabular}{p{1.0cm}p{6.0cm}|p{1.1cm}p{6.0cm}}
\hline
BT-S & es ist eine Nachricht , die die aktive Substanz {\bf enth\"alt} . & BT-T & Invirase is a {\bf medicine} containing the active substance saquinavir . \\ \hline\hline
% BT-S2 & Buchung ist eine Technologie , die die aktive Substanz {\bf enth\"alt} . & BT-T2 & Actos is a medicine containing the active substance pioglitazone . \\ \hline\hline
Test-S & ABILIFY ist ein {\bf Arzneimittel} , das den Wirkstoff Aripiprazol {\bf enth\"alt} . & Test-T & Prevenar is a {\bf medicine} containing the design of Arixtra .   \\ \hline
% \multicolumn{3}{}{} Output & Prevenar is a medicine containing the design of Arixtra .              \\ \hline
\end{tabular} }
% \vspace{-1mm}
\caption{An example that shows why BT could translate the OOV word ``Arzneimittel'' correctly into ``medicine''. ``enth\'alt'' corresponds to the English word ``contain''. Though BT can't translate a correct source sentence for augmentation, it generates sentences with certain patterns that could be identified by the model, which helps translate in-domain unseen words.}
\label{tab:bt-ex}
% \vspace{-3mm}
\end{table*}

\subsection{When do Copy, BT and DALI Work?}
From Figure~\ref{fig:word-level-1}, we can see that Copy, BT and DALI  all improve the translation accuracy of in-domain unseen words. In this section, we explore exactly what types of words each method improves on. We randomly pick some in-domain unseen word pairs which are translated 100\% correctly in the translation outputs of systems trained with each method. We also count these word pairs' occurrences in the pseudo-in-domain training set. The examples are demonstrated in Table~\ref{tab:word-translation}. 

We find that in the case of Copy, over $80\%$ of the successful word translation pairs have the same spelling format for both source and target words, and almost all of the rest of the pairs share subword components.
In short, and as expected, Copy excels on improving accuracy of words that have identical forms on the source and target sides.

As expected, our proposed method mainly increases the translation accuracy of the pairs in our induced lexicon.
It also leverages the subword components to successfully translate compound words. For example, ``monotherapie'' does not occur in our induced lexicon, but the model is still able to translate it correctly based on its subwords ``mono@@'' and ``therapie'' by leveraging the successfully induced pair ``therapie'' and ``therapy''.

\begin{table}[t]
\resizebox{0.48\textwidth}{!}{
\begin{tabular}{l|l|r}
\hline
Type & Word Pair & Count          \\ \hline\hline
Copy & (tremor, tremor) & 452 \\
& (347, 347) & 18   \\ \hline
BT   & (ausschuss, committee) & 0 \\
& (apotheker, pharmacist) & 0 \\ 
& (toxizit\"{a}t, toxicity) & 0 \\ \hline
DALI & (m\"{u}digkeit, tiredness) & 444 \\
& (therapie, therapy)& 9535 \\ 
& (monotherapie, monotherapy)& 0         
\\ \hline
\end{tabular} }
% \vspace{-1mm}
\caption{100\% successful word translation examples from the output of the IT to Medical adaptation task. The Count column shows the number of occurrences of word pairs in the pseudo-in-domain training set.}
\label{tab:word-translation}
% \vspace{-3mm}
\end{table}% 

% \begin{table*}[th]
% \begin{tabular}{|l|l|}
% \hline
% Train Src 1 & der ase empfohlen , die von dort vorgegebenen Namen gegeben wurden .                      \\ \hline
% Train Tgt 1 & the Committee recommended that Raptiva be given marketing authorisation .                 \\ \hline
% % Train Src 2 & der Sprungziel empfohlen , dass sie mit eigenen Schnittangeschlossen .                    \\ \hline
% % Train Tgt 2 & the Committee recommended that MicardisPlus be given marketing authorisation .            \\ \hline
% Test Src    & Der Ausschuss empfahl , die Genehmigung für das Inverkehrbringenvon Abilify zu erteilen . \\ \hline
% Test Tgt    & the Committee recommended that Abilify be given marketing authorisation .                 \\ \hline
% Output      & The Committee responsible for the production of clinical studies is demonstrated .        \\ \hline
% \end{tabular}
% \end{table*}

% \gn{This paragraph seemed really long for something relatively unimportant. I think it coukld be cut down significantly.}
It is more surprising to find that adding a back translated corpus significantly improves the model's ability to translate in-domain unseen words correctly, even if the source word never occurs in the pseudo-in-domain corpus. Even more surprisingly, we find that the majority of the correctly translated source words are not segmented at all, which means that the model does not leverage the subword components to make correct translations. In fact, for most of the correctly translated in-domain word pairs, the source words are never seen during training. To further analyze this, we use our BT model to do word-for-word translation for these individual words without any other context, and the results turn out to be extremely bad, indicating that the model does not actually find the correspondence of these word pairs. Rather, it rely solely on the decoder to make the correct translation on the target side for test sentences with related target sentences in the training set. To verify this, Table~\ref{tab:bt-ex} demonstrates an example extracted from the pseudo-in-domain training set. BT-T shows a monolingual in-domain target sentence and BT-S is the back-translated source sentence. Though the back translation fails to generate any in-domain words and the meaning is unfaithful, it succeeds to generate a similar sentence pattern as the correct source sentence, which is ``... ist eine (ein) ... , die (das) ... enth\"alt .''. The model can easily detect the pattern through the attention mechanism and translate the highly related word ``medicine'' correctly.

% Also, the word-for-word translation breaks the tie of the originally highly related sentences, which makes it unable to translate words merely based on contexts.

From the above analysis, it can be seen that the improvement brought by the augmentation of BT and DALI are largely orthogonal. The former utilizes the highly related contexts to translate unseen in-domain words while the latter directly injects reliable word translation pairs to the training corpus. This explains why we get further improvements over either single method alone.

\subsection{Lexicon Coverage}
\label{sec:lex_coverage}
\begin{figure}
  \includegraphics[width=1.06\linewidth]{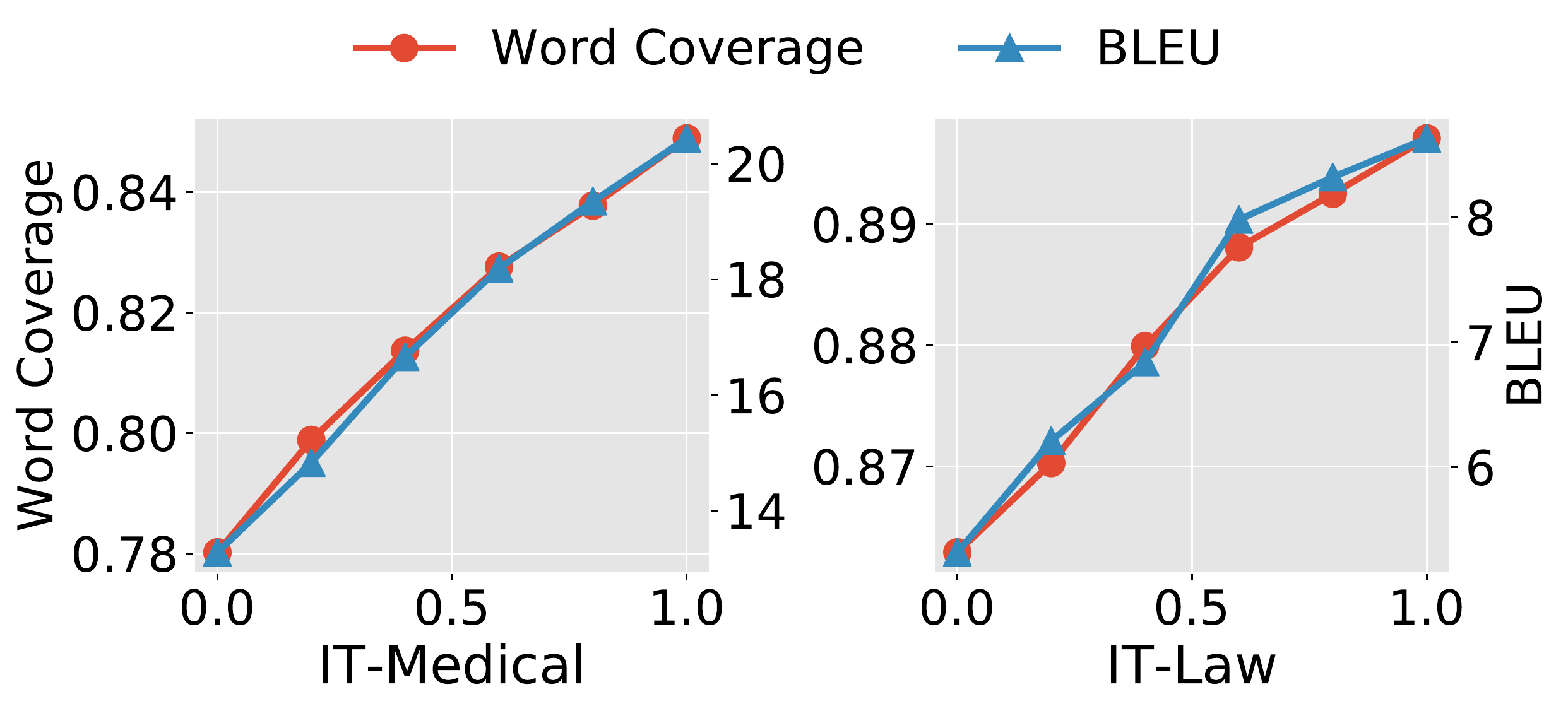}
%   \vspace{-6mm}
  \caption{Word coverage and BLEU score of the Medical test set when the pseudo-in-domain training set is constructed with different level of lexicon coverage.}
  \label{fig:word-cover}
%   \vspace{-3mm}
\end{figure}
Intuitively, with a larger lexicon, we would expect a better adaptation performance. In order to examine this hypothesis, we do experiments using pseudo-in-domain training sets generated by our induced lexicon with various coverage levels. Specifically, we split the lexicon into 5 folds randomly and use a portion of it comprising folds 1 through 5, which correspond to 20\%, 40\%, 60\%, 80\% and 100\% of the original data. We calculate the coverage of the words in the Medical test set comparing with each pseudo-in-domain training set. We use each training set to train a model and get its corresponding BLEU score. From Figure~\ref{fig:word-cover}, we find that the proportion of the used lexicon is highly correlated with both the known word coverage in the test set and its BLEU score, indicating that by inducing a larger and more accurate lexicon, further improvements can likely be made.

% \subsection{Supervised vs Unsupervised Lexicons vs GIZA++ Lexicons}
% Since we use both supervised and unsupervised methods to induce the lexicon, we would like to examine.
\begin{table*}[t]
\centering
\resizebox{0.9\textwidth}{!}{
\begin{tabular}{l||l||l}
\hline
Source    & ABILIFY ist ein Arzneimittel , das den Wirkstoff Aripiprazol enthält .  & BLEU  \\ 
Reference    & abilify is a medicine containing the active substance aripiprazole .   & 1.000  \\  \hline
Unadapted   & the time is a figure that corresponds to the formula of a formula .     & 0.204 \\  
Copy   & abilify is a casular and the raw piprexpression offers .                & 0.334 \\ 
BT     & prevenar is a medicine containing the design of arixtra .                & 0.524\\ 
DALI    & abilify is a arzneimittel that corresponds to the substance ariprazole . & 0.588 \\
DALI+BT & abilify is a arzneimittel , which contains the substance aripiprazole .  & 0.693\\ \hline
\end{tabular} }
\caption{Translation outputs from various data augmentation method and our method for IT$\rightarrow$Medical adaptation.}
% \vspace{-3mm}
\end{table*}

% {\color{red} 
\subsection{Semi-supervised Adaptation}
Although we target \textit{unsupervised} domain adaptation, it is also common to have a limited amount of in-domain parallel sentences in a semi-supervised adaptation setting. To measure efficacy of DALI in this setting, we first pre-train an NMT model on a parallel corpus in the IT domain, and adapt it to the medical domain. The pre-trained NMT obtains 7.43 BLEU scores on the medical test set. During fine-tuning, we sample 330,278 out-of-domain parallel sentences, and concatenate them with 547,325 pseudo-in-domain sentences generated by DALI and the real in-domain sentences. We also compare the performance of fine-tuning on the combination of the out-of-domain parallel sentences with only real in-domain sentences. We vary the number of real in-domain sentences in the range of [20K, 40K, 80K, 160K, 320K, 480K]. In Figure~\ref{fig:semi}, semi-supervised adaptation outperforms unsupervised adaptation after we add more than 20K real in-domain sentences. As the number of real in-domain sentences increases, the BLEU scores on the in-domain test set improve, and fine-tuning on both the pseudo and real in-domain sentences further improves over fine-tuning sorely on the real in-domain sentences. In other words, given a reasonable number of real in-domain sentences in a common semi-supervised adaptation setting, DALI is still helpful in leveraging a large number of monolingual in-domain sentences.

\subsection{Effect of Out-of-Domain Corpus}

\begin{figure}%
\centering
\subfigure[IT-Medical]{%
\label{fig:semi}%
\includegraphics[width=0.25\textwidth]{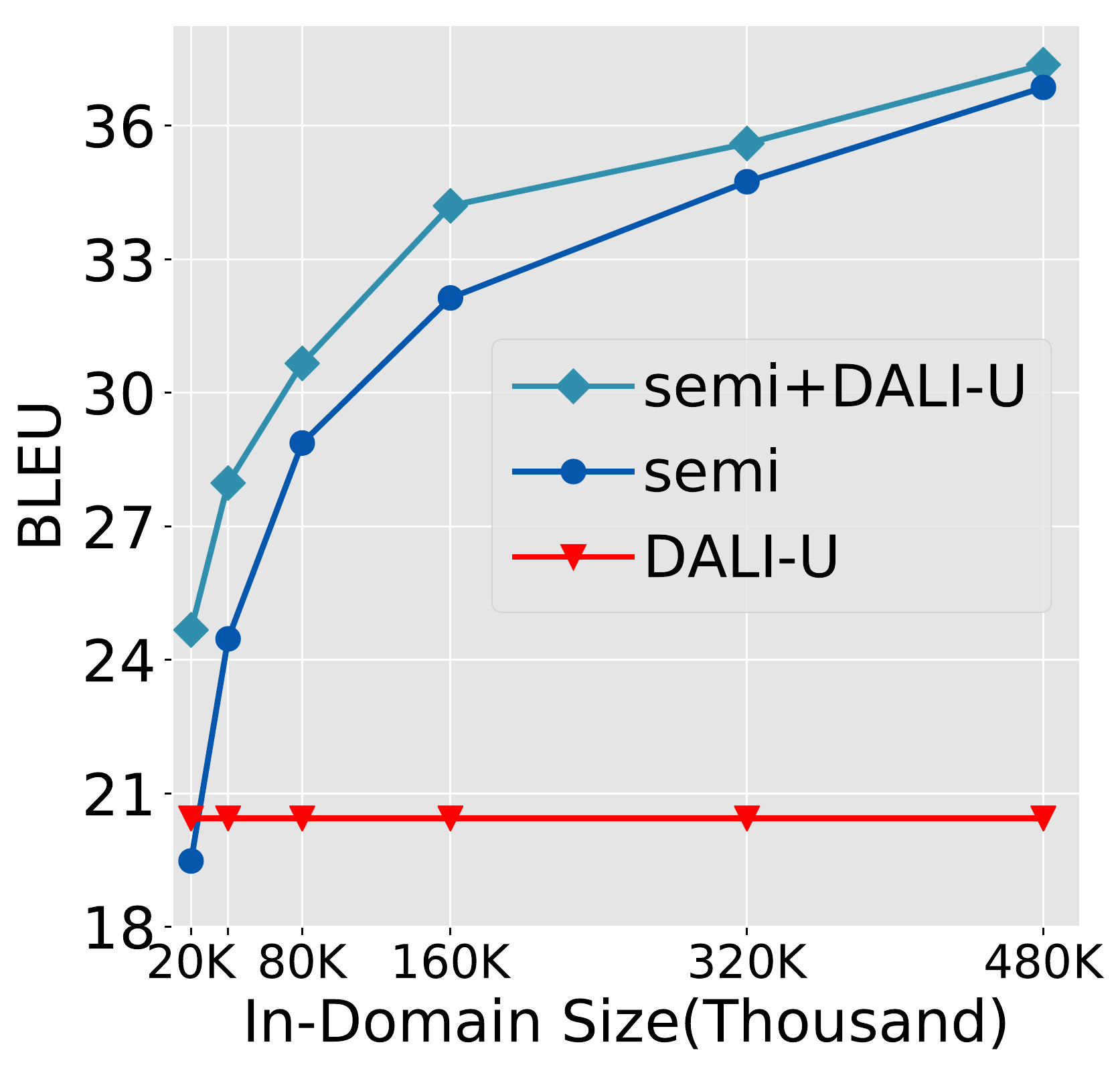}}%
\subfigure[Subtitles-Medical]{%
\label{fig:out-size}%
\includegraphics[width=0.25\textwidth]{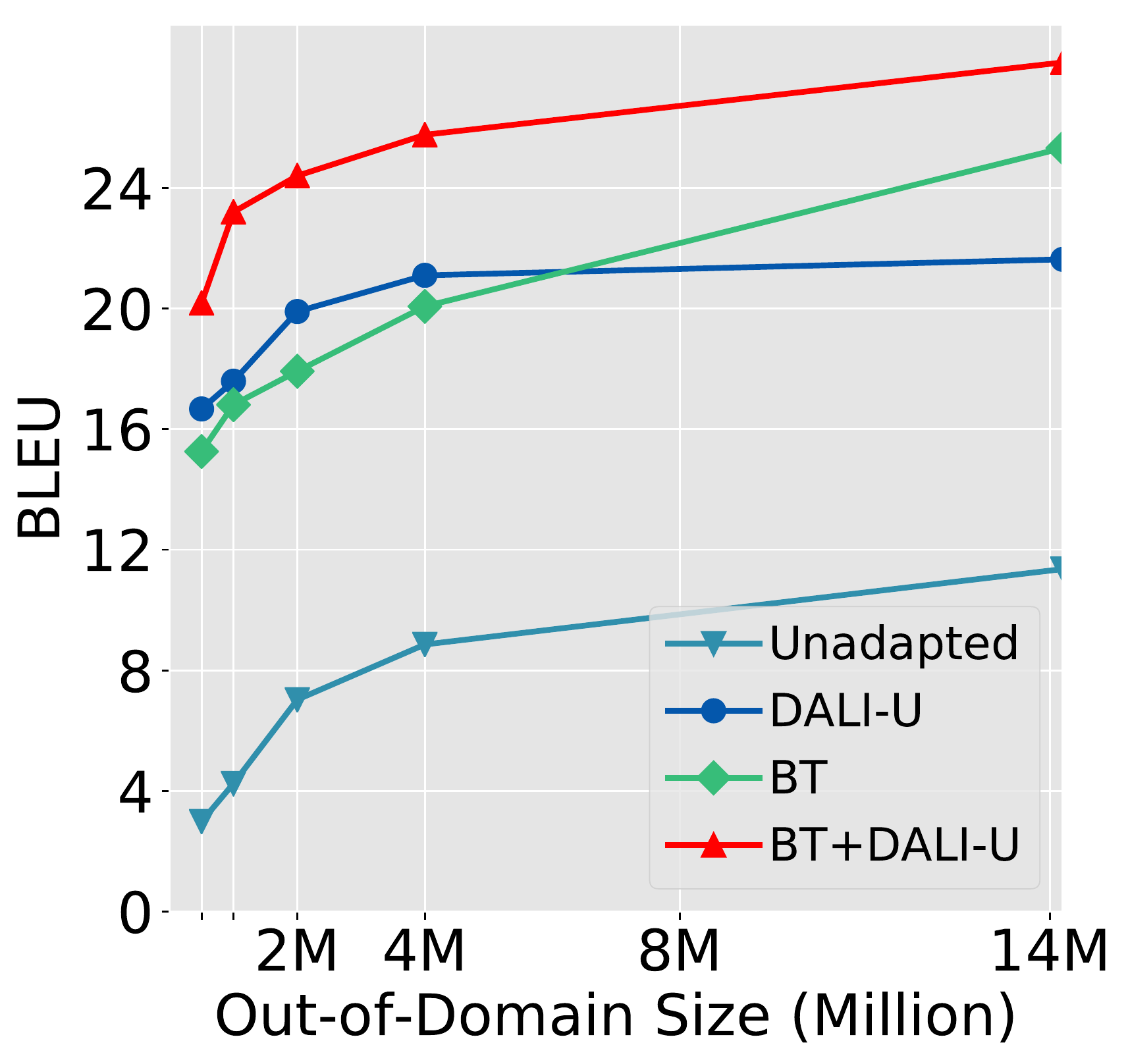}}%
% \vspace{-4mm}
\caption{Effect of training on increasing number of in-domain (a) and out-of-domain (b) parallel sentences}
% \vspace{-3mm}
\end{figure}

The size of data that we use to train the unadapted NMT and BT NMT models varies from hundreds of thousands to millions, and covers a wide range of popular domains. Nonetheless, the unadapted NMT and BT NMT models can both benefit from training on a large out-of-domain corpus. We examine the question: how does fine-tuning on weak and strong unadapted NMT models affect the adaptation performance? To this end, we compare DALI and BT on adapting from subtitles to medical domains, where the two largest corpus in subtitles and medical domains have 13.9 and 1.3 million sentences. We vary the size of out-of-domain corpus in a range of $[0.5,1,2,4,13.9]$ million, and fix the number of in-domain target sentences to 0.6 million. In Figure~\ref{fig:out-size}, as the size of out-of-domain parallel sentences increases, we have a stronger upadapted NMT which consistently improves the BLEU score of the in-domain test set. Both DALI and BT also benefit from adapting a stronger NMT model to the new domain. Combining DALI with BT further improves the performance, which again confirms our finding that the gains from DALI and BT are orthogonal to each other. Having a stronger BT model improves the quality of synthetic data, while DALI aims at improving the translation accuracy of OOV words by explicitly injecting their translations.

\subsection{Effect of Domain Coverage}
We further test the adaptation performance of DALI when we train our base NMT model on the WMT14 German-English parallel corpus. The corpus is a combination of Europarl v7, Common Crawl corpus and News Commentary, and consists of 4,520,620 parallel sentences from a wider range of domains. In Table~\ref{tab:wmt-adapt}, we compare the BLEU scores of the test sets between the unadapted NMT and the adapted NMT using DALI-U. We also show the percentage of source words or subwords in the training corpus of five domains being covered by the WMT14 corpus. Although the unadapted NMT system trained on the WMT14 corpus obtains higher scores than that trained on the corpus of each individual domain, DALI still improves the adaptation performance over the unadapted NMT system by up to 5 BLEU score.  
\begin{table}[]
\centering
\resizebox{0.49\textwidth}{!}{
\begin{tabular}{l|r|r|r|r}
\hline
Domain & Base & DALI & Word  & Subword \\ \hline
Medical & 28.94 & 30.06 & 44.1\% & 69.1\% \\ 
IT & 18.27 & 23.88 & 45.1\% & 77.4\% \\ 
Subtitles & 22.59 & 22.71 & 35.9\% & 62.5\% \\ 
Law & 24.26 & 24.55 & 59.0\% & 73.7\% \\ 
Koran & 11.64 & 12.19 & 83.1\% & 74.5\% \\ \hline
\end{tabular}}
\caption{BLEU scores of LSTM based NMT models when trained on WMT14 De-En data (Base), and adapted to one domain (DALI). The last two columns show the percentage of source word/subword overlap between the training data on the WMT domain and other five domains.}
\label{tab:wmt-adapt}
% \vspace{-3mm}
\end{table}

% }

\subsection{Qualitative Examples}
Finally, we show outputs generated by various data augmentation methods. Starting with the unadapted output, we can see that the output is totally unrelated with the reference. By adding the copied corpus, words that have the same spelling in the source and target languages e.g. ``abilify'' are correctly translated. With back translation, the output is more fluent; though keywords like ``abilify'' are not well translated, in-domain words that are highly related with the context like ``medicine'' are correctly translated. DALI manages to translate in-domain words like ``abilify'' and ``substance'', which are added by DALI using the induced lexicon. By combining both BT and DALI, the output becomes fluent and also contains correctly translated in-domain keywords of the sentence.

\section{Related Work}
There is much work on \emph{supervised} domain adaptation setting where we have large out-of-domain parallel data and much smaller in-domain parallel data. \newcite{luong2015stanford} propose training a model on an out-of-domain corpus and do fine-tuning with small sized in-domain parallel data to mitigate the domain shift problem. Instead of naively mixing out-of-domain and in-domain data,~\newcite{W17-4712} circumvent the domain shift problem by jointly learning domain discrimination and the translation. \newcite{D15-1147} and \newcite{wang-EtAl:2017:Short3} address the domain adaptation problem by assigning higher weight to out-of-domain parallel sentences that are close to the in-domain corpus. Our proposed method focuses on solving the adaptation problem with no in-domain parallel sentences, a strict unsupervised setting. 

%\gn{This sounds repetitive after we've already compared to back-translation and data copying extensively before this section. Try to remove the stuff you've already said and focus on the new stuff.}
Prior work on using monolingual data to do data augmentation could be easily adapted to the domain adaptation setting. Early studies on data-based methods such as self-enhancing~\cite{schwenk2008investigations,lambert-etal-2011-investigations} translate monolingual source sentences by a statistical machine translation system, and continue training the system on the synthetic parallel data. Recent data-based methods such as back-translation~\cite{sennrich-haddow-birch:2016:P16-11} and copy-based methods~\cite{currey-micelibarone-heafield:2017:WMT} mainly focus on improving fluency of the output sentences and translation of identical words, while our method targets OOV word translation. In addition, there have been several attempts to do data augmentation using monolingual source sentences~\cite{zhang-zong:2016:EMNLP2016,chinearios-peris-casacuberta:2017:WMT}. Besides, model-based methods change model architectures to leverage monolingual corpus by introducing an extra learning objective, such as auto-encoder objective~\cite{cheng-EtAl:2016:P16-1} and language modeling objective~\cite{ramachandran-liu-le:2017:EMNLP2017}. Another line of research on using monolingual data is unsupervised machine translation~\cite{artetxe2017unsupervised, lample-EtAl:2018:EMNLP, lample2017unsupervised,yang-EtAl:2018:Long}. These methods use word-for-word translation as a component, but require a careful design of model architectures, and do not explicitly tackle the domain adaptation problem. Our proposed data-based method does not depend on model architectures, which makes it orthogonal to these model-based methods. 

%share similar methodology as our data augmentation method of word-for-word translation, but our goals are explicitly different. These two methods 
% Semi-supervised NMT~\cite{cheng-EtAl:2016:P16-1} introduces an autoencoder objective to learn source-to-source and target-to-target translations using monolingual sentences. The method is model-based and therefore is orthogonal to our data-based method. Unsupervised pre-training method~\cite{ramachandran-liu-le:2017:EMNLP2017} initializes NMT's encoder and decoder by two language models pre-trained on monolingual corpus. This pre-training technique is also orthogonal to our method. 
%Back-translation method~\cite{sennrich-haddow-birch:2016:P16-11} augments the original training corpus by translating monolingual target sentences into source sentences and construct pseudo parallel sentences.  Recent results in~\cite{currey-micelibarone-heafield:2017:WMT} show that simply copying the target sentences to the source side and training the NMT system on the augmented corpus can obtain significant improvements for low-resource NMT. 

Our work shows that apart from strengthening the target-side decoder, direct supervision over the in-domain unseen words is essential for domain adaptation. Similar to this, a variety of methods focus on solving OOV problems in translation. \newcite{daume2011domain} induce lexicons for unseen words and construct phrase tables for statistical machine translation. However, it is nontrivial to integrate lexicon into NMT models that lack explicit use of phrase tables. With regard to NMT, ~\newcite{arthur-neubig-nakamura:2016:EMNLP2016} use a lexicon to bias the probability of the NMT system and show promising improvements.~\newcite{luong2015stanford} propose to emit OOV target words by their corresponding source words and do post-translation for those OOV words with a dictionary. ~\newcite{fadaee-bisazza-monz:2017:Short2} propose an effective data augmentation method that generates sentence pairs containing rare words in synthetically created contexts, but this requires parallel training data not available in the fully unsupervised adaptation setting. ~\newcite{Arcan2017TranslatingDE} leverage a domain-specific lexicon to replace unknown words after decoding. ~\newcite{zhao-etal-2018-addressing} design a contextual memory module in an NMT system to memorize translations of rare words. ~\newcite{kothur-etal-2018-document} treats an annotated lexicon as parallel sentences and continues training the NMT system on the lexicon. Though all these works leverage a lexicon to address the problem of OOV words, none specifically target translating in-domain OOV words under a domain adaptation setting.

% \gn{Maybe also briefly mention unsupervised translation, which uses word-by-word translation to bootstrap?}

% \jh{list of paper on lexicons and OOV words}

% \begin{enumerate}
%     \item SMT that learns unseen words using lexicon induction~\cite{daume2011domain}
%     \item Replace rare words by UNK, and translate UNK by the lexicon after translation~\cite{luong-EtAl}. 
%     \item Interpolation of LM probability~\cite{domhan-hieber:2017:EMNLP2017}
%   \item data augmentation using lexicon~\cite{fadaee-bisazza-monz:2017:Short2}
%   \item sub-word units to address OOV words~\cite{sennrich-haddow-birch:2016:P16-12}
% \end{enumerate}

% \jh{work on leveraging monolingual data}

% \jh{Domain Adaptation for NMT}

% \begin{enumerate}
%     \item Mixed parallel sentences in both domains and fine-tune the NMT~\cite{chu-dabre-kurohashi:2017:Short}
%     \item Continue training on the in-domain parallel data~\cite{luong2015stanford,freitag2016fast}
%     \item Evaluation metric for domain adaptation~\cite{etchegoyhen2018evaluating}
%     \item Jointly learn domain discrimination and translation~\cite{W17-4712}
%     \item \cite{wang-EtAl:2017:EMNLP20174}
% \end{enumerate}

\section{Conclusion}
\label{sec:conclusion}
In this paper, we propose a \emph{data-based, unsupervised adaptation} method that focuses on domain adaption by lexicon induction (DALI) for mitigating unknown word problems in NMT. We conduct extensive experiments to show consistent improvements of two popular NMT models through the usage of our proposed method. Further analysis show that our method is effective in fine-tuning a pre-trained NMT model to correctly translate unknown words when switching to new domains.

\section*{Acknowledgements}
The authors thank anonymous reviewers for their constructive comments on this paper. This material is based upon work supported by the Defense Advanced Research Projects Agency Information Innovation Office (I2O) Low Resource
Languages for Emergent Incidents (LORELEI)
program under Contract No. HR0011-15-C0114.
The views and conclusions contained in this document are those of the authors and should not be
interpreted as representing the official policies, either expressed or implied, of the U.S. Government. The U.S. Government is authorized to reproduce and distribute reprints for Government
purposes notwithstanding any copyright notation
here on.

\bibliography{myabbrv,main.bib}
\bibliographystyle{acl_natbib}

\cleardoublepage
\appendix

\section{Appendices}
\label{sec:appendix}

% \subsection{Data Statistic} \label{sec:data_stat}
% \begin{table}[h]
% \centering
% \begin{tabular}{l||r||r||r}
% \hline
% Corpus         & Words     & Sentences & W/S \\ \hline
% Medical & 12,867,326  & 1,094,667   & 11.76     \\ 
% IT             & 2,777,136   & 333,745    & 8.32  \\
% Subtitles      & 106,919,386 & 13,869,396  & 7.71  \\
% Law  & 15,417,835  & 707,630    & 21.80     \\ 
% Koran & 9,598,717   & 478,721    & 20.05     \\ \hline
% \end{tabular}
% \caption{Corpus statistics over five domains.}
% \label{tab:stats}
% \end{table}

\subsection{Hyper-parameters}
For the RNN-based model, we use two stacked LSTM layers for both the encoder and the decoder with a hidden size and a embedding size of 512, and use feed-forward attention~\cite{bahdanau2014neural}. We use a Transformer model building on top of the OpenNMT toolkit~\cite{opennmt} with six stacked self-attention layers, and a hidden size and a embedding size of 512. The learning rate is varied over the course of training~\cite{vaswani2017attention}.
\begin{table}[h]
\resizebox{0.48\textwidth}{!}{
\begin{tabular}{l|l|l}
\hline
                        & LSTM  & XFMR \\ \hline
Embedding size          & 512            & 512         \\ 
Hidden size             & 512            & 512         \\ 
\# encoder layers   & 2              & 6           \\ 
\# decoder layers   & 2              & 6           \\ \hline
Batch                   & 64 sentences   & 8096 tokens \\ 
Learning rate           & 0.001          & -     \\ 
Optimizer               & Adam           & Adam       \\ \hline
Beam size               & 5              & 5           \\ 
Max decode length & 100            & 100         \\ \hline
\end{tabular}}
\caption{Configurations of LSTM-based NMT and Transformer (XFMR) NMT, and tuning parameters during training and decoding}
\label{tab:parameters}
\end{table}

\subsection{Domain Shift}
To measure the extend of domain shift, we train a 5-gram language model on the target sentences of the training set on one domain, and compute the average perplexity of the target sentences of the training set on the other domain. In Table~\ref{tab:ppl}, we can find significant differences of the average perplexity across domains.
\begin{table}[th]
\resizebox{0.48\textwidth}{!}{
\begin{tabular}{l|l|l|l|l|l}
\hline
Domain   & Medical & IT   & Subtitles & Law  & Koran \\ \hline\hline
Medical   & 1.10    & 2.13 & 2.34      & 1.70 & 2.15  \\ \hline
IT        & 1.95    & 1.21 & 2.06      & 1.83 & 2.05  \\ \hline
Subtitles & 1.98    & 2.13 & 1.31      & 1.84 & 1.82  \\ \hline
Law       & 1.88    & 2.15 & 2.50      & 1.12 & 2.16  \\ \hline
Koran     & 2.09    & 2.23 & 2.08      & 1.94 & 1.11  \\ \hline
\end{tabular}}
\caption{Perplexity of 5-gram language model trained on one domain (columns) and tested on another domain (rows)}
\label{tab:ppl}
\end{table}

\subsection{Lexicon Overlap}
Table~\ref{tab:lex_coverage} shows the overlap of the induced lexicons from supervised, unsupervised induction and GIZA++ extraction across five domains. The second and third column show the percentage of unique lexicons induced only by unsupervised induction and supervised induction respectively, while the last column shows the percentage of the lexicons induced by both methods.
\begin{table}[th]
\centering
\resizebox{0.48\textwidth}{!}{
\begin{tabular}{l||c|c||c}
\hline
Corpus    & Unsupervised & Supervised   & Intersection \\ \hline\hline
Medical      & 5.3\% & 5.4\% & 44.7\%       \\
IT        & 4.1\% & 4.1\% & 45.2\%       \\
Subtitles & 1.0\% & 1.0\% & 37.1\%       \\
Law    & 4.4\% & 4.5\% & 45.7\%       \\
Koran     & 2.1\% & 2.0\% & 40.6\%       \\ \hline
\end{tabular}}
% \vspace{-0.5em}
\caption{Lexicon overlap between supervised, unsupervised and GIZA++ lexicon.}
\label{tab:lex_coverage}
% \vspace{-2.3em}
\end{table}

% \subsection{Out-of-Vocabulary Statistics}
% We show the OOV statistics across five domains in Table~\ref{tab:oov_statistic} adn \ref{tab:en_oov}. 
\begin{table*}[t]
    \centering
    \begin{tabular}{c|r|r|r|r|r|r} \hline 
Domain & $|$In$|$ & Medical & IT & Subtitles & Law & Koran \\\hline 
Medical & 125724 &  0 (0.00) &  123670 (0.98) &  816762 (6.50) &  159930 (1.27) &  12697 (0.10) \\\hline 
IT & 140515 &  108879 (0.77) &  0 (0.00) &  818303 (5.82) &  167630 (1.19) &  12512 (0.09) \\\hline 
Subtitles & 857527 &  84959 (0.10) &  101291 (0.12) &  0 (0.00) &  129323 (0.15) &  3345 (0.00) \\\hline 
Law & 189575 &  96079 (0.51) &  118570 (0.63) &  797275 (4.21) &  0 (0.00) &  10899 (0.06) \\\hline 
Koran & 18292 &  120129 (6.57) &  134735 (7.37) &  842580 (46.06) &  182182 (9.96) &  0 (0.00) \\\hline
    \end{tabular}
    \caption{Out-of-Vocabulary statistics of German Words across five domains. Each row indicates the OOV statistics of the out-of-domain (row) corpus against the in-domain (columns) corpus. The second column shows the vocabulary size of the out-of-domain corpus in each row. The remaining columns (3rd-7th) show the number of domain-specific words in each in-domain corpus with respect to the out-of-domain corpus, and the ratio between the number of out-of-domain corpus and the domain specific words. }
    \label{tab:oov_statistic}
\end{table*}

\begin{table*}[t]
    \centering
    \begin{tabular}{c|r|r|r|r|r|r} \hline 
Domain & $|$In$|$ & Medical & IT & Subtitles & Law & Koran \\\hline 
Medical & 68965 &  0 (0.00) &  57206 (0.83) &  452166 (6.56) &  72867 (1.06) &  15669 (0.23) \\\hline 
IT & 70652 &  55519 (0.79) &  0 (0.00) &  448072 (6.34) &  75318 (1.07) &  14771 (0.21) \\\hline 
Subtitles & 480092 &  41039 (0.09) &  38632 (0.08) &  0 (0.00) &  53984 (0.11) &  4953 (0.01) \\\hline 
Law & 92501 &  49331 (0.53) &  53469 (0.58) &  441575 (4.77) &  0 (0.00) &  13399 (0.14) \\\hline 
Koran & 22450 &  62184 (2.77) &  62973 (2.81) &  462595 (20.61) &  83450 (3.72) &  0 (0.00) \\\hline 
    \end{tabular}
    \caption{Out-of-Vocabulary statistics of English Words across five domains. Each row indicates the OOV statistics of the out-of-domain (row) corpus against the in-domain (columns) corpus. The second column shows the vocabulary size of the out-of-domain corpus in each row. The remaining columns (3rd-7th) show the number of domain-specific words in each in-domain corpus with respect to the out-of-domain corpus, and the ratio between the number of out-of-domain corpus and the domain specific words. }
    \label{tab:en_oov}
\end{table*} 

\end{document}